\pdfoutput=1

\documentclass[11pt]{article}

\usepackage{acl}

\usepackage{times}
\usepackage{latexsym}
\usepackage{booktabs}
\usepackage{graphicx}
\usepackage{multirow}
\usepackage{times}
\usepackage{amsfonts,amsmath,amssymb,bm}
\usepackage{graphicx}
\usepackage{xcolor}
\usepackage{setspace}
\usepackage{pdflscape}
\usepackage{pdfpages}
\usepackage{enumitem}
\usepackage{rotating}
\usepackage{caption}
\usepackage{subcaption}

\DeclareMathOperator*{\argmax}{arg\,max}

\usepackage[english]{babel}
\usepackage{hyperref}
\addto\captionsenglish{%
}
\addto\extrasenglish{%
}

\usepackage[T1]{fontenc}

\usepackage[utf8]{inputenc}

\usepackage{microtype}
\usepackage{setspace}

\usepackage{inconsolata}

\title{Generative Language Models for Paragraph-Level Question Generation}

\author{Asahi Ushio \and Fernando Alva-Manchego \and Jose Camacho-Collados\\
  Cardiff NLP, School of Computer Science and Informatics, Cardiff University, UK\\
  \texttt{\{UshioA,AlvaManchegoF,CamachoColladosJ\}@cardiff.ac.uk}
  \\}

\begin{document}
\maketitle


\begin{abstract}
Powerful generative models have led to recent progress in \textit{question generation} (QG). However, it is difficult to measure advances in QG research since there are no standardized resources that allow a uniform comparison among approaches. In this paper, we introduce QG-Bench, a multilingual and multidomain benchmark for QG that unifies existing question answering datasets by converting them to a standard QG setting. It includes general-purpose datasets such as SQuAD \cite{rajpurkar-etal-2016-squad} for English, datasets from ten domains and two styles, as well as datasets in eight different languages. Using QG-Bench as a reference, we perform an extensive analysis of the capabilities of language models for the task. First, we propose robust QG baselines based on fine-tuning generative language models. Then, we complement automatic evaluation based on standard metrics with an extensive manual evaluation, which in turn sheds light on the difficulty of evaluating QG models. Finally, we analyse both the domain adaptability of these models as well as the effectiveness of multilingual models in languages other than English.
QG-Bench is released along with the fine-tuned models presented in the paper,\footnote{\url{https://github.com/asahi417/lm-question-generation}} which are also available as a demo.\footnote{\url{https://autoqg.net/}}

\end{abstract}

\section{Introduction}
\begin{figure}[t]
    \centering
    \includegraphics[width=0.95\columnwidth]{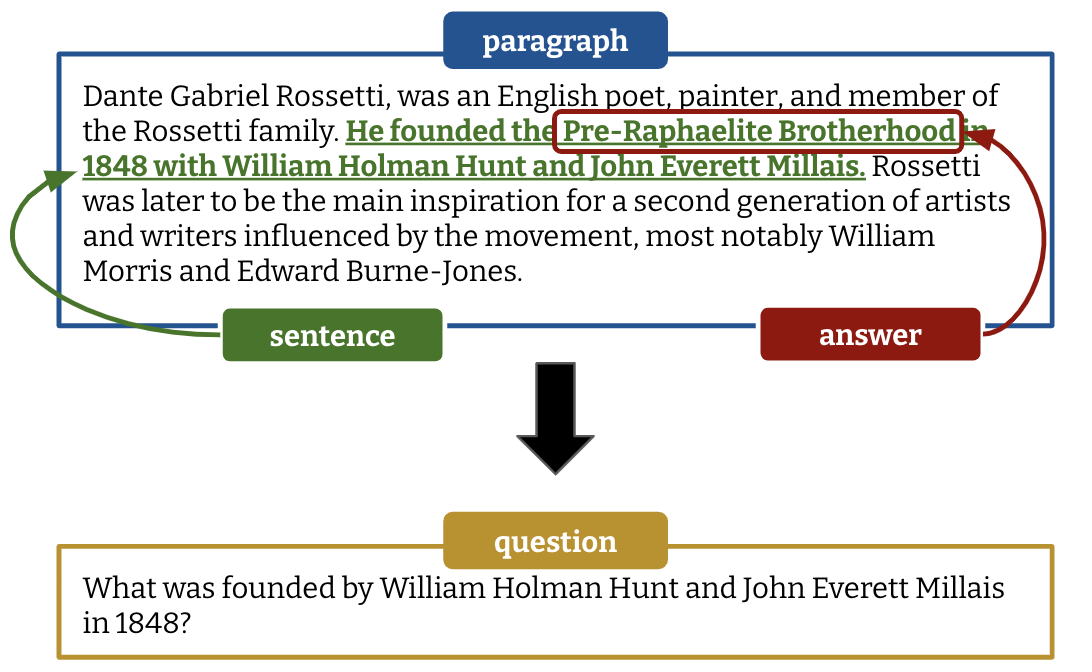}
    \caption{Overview of paragraph-level QG.}
    \label{fig:task_overview}
\end{figure}

Question generation \citep[QG,][]{mitkov-ha-2003-computer} is the task of generating a question given an input context consisting of a document, a paragraph or a sentence, and an answer where the question is anchored (see \autoref{fig:task_overview}). 
QG has been widely studied in natural language processing communities \cite{du-etal-2017-learning,zhou2017neural,du-cardie-2018-harvesting}, and it has recently been exploited to train question answering (QA) models without human supervision \cite{lewis-etal-2019-unsupervised,zhang-bansal-2019-addressing,puri-etal-2020-training}, or as a means of data augmentation \cite{shakeri-etal-2020-end,bartolo-etal-2021-improving}. 
It has also been applied to develop educational systems \cite{heilman-smith-2010-good,lindberg-etal-2013-generating}, information retrieval models \cite{pyatkin-etal-2021-asking,lewis-etal-2021-paq}, and for model interpretation \cite{perez-etal-2020-unsupervised,lee-etal-2020-generating}.

Despite its success in downstream applications, the development of neural QG models has received less attention. 
For example, the choice of the base pre-trained model is arbitrary (without proper justification in most cases) as it is not straightforward to compare different models.
As a consequence, while ERNIE-GEN \cite{ERNIE-GEN} and UniLMv2 \cite{UNILMv2} are current SotA in the SQuAD QG benchmark \cite{du-etal-2017-learning}, 
T5 \cite{T5} and BART \cite{lewis-etal-2020-bart} are used in many applications in practice 
\cite{paranjape2021retrieval,bartolo-etal-2021-improving,lewis-etal-2021-paq,pyatkin-etal-2021-asking}. 

A possible reason is inconsistent evaluation and comparison of QG models, due to the lack of appropriate evaluation protocols and benchmarks. 
For instance, evaluation of QG models relies on BLEU4 \cite{papineni-etal-2002-bleu}, METEOR \cite{denkowski-lavie-2014-meteor}, and ROUGE\textsubscript{L} \cite{lin-2004-rouge}, with human-made questions as references. However, some of these metrics may have low correlation with human judgements, especially when it comes to \emph{answerability}, since they tend not to take the associated answer into account \cite{nema-khapra-2018-towards}.
Moreover, QG applications can use different contexts as input, such as sentence-level \cite{pyatkin-etal-2021-asking,lewis-etal-2019-unsupervised} vs paragraph-level \cite{zhang-bansal-2019-addressing,puri-etal-2020-training}, or answer-aware \cite{shakeri-etal-2020-end,bartolo-etal-2021-improving} vs answer-free \cite{lopez2020transformer}. 
These are generally used interchangeably in the literature. 

To investigate how to tackle the issues previously raised, we introduce QG-Bench, a collection of standard QA datasets unified into a single benchmark, including domain-specific datasets and for eight different languages (\autoref{sec:benchmark}).
We then use QG-Bench to fine-tune various generative language models (LMs) by formulating paragraph-level QG as a sequence-to-sequence generation task (\autoref{sec:qg-models}), and measure their performance on in-domain and language-specific data (\autoref{sec:autoeval}). 
Finally, we present a multi-faceted analysis of our QG models by varying their input context size (\autoref{sec:anlysis_model_input}), conducting a manual evaluation (\autoref{sec:result_manual_evaluation}), and studying their abilities for domain adaptation (\autoref{sec:domainadapt}).

\section{Related Work}
Early work on QG was based on human-engineered templates \cite{mitkov-ha-2003-computer,rus-etal-2010-first} and well-designed pipelines \cite{heilman-smith-2010-good,labutov-etal-2015-deep}, 
but soon neural approaches took over by generating a question from a text in an end-to-end manner \cite{du-etal-2017-learning,zhou2017neural,du-cardie-2018-harvesting}.
The quality of QG models was later improved by masked LM pre-training \cite{devlin-etal-2019-bert,RoBERTa} where the encoder of the QG model is fine-tuned from pre-trained LMs \cite{chan-fan-2019-recurrent,zhang-bansal-2019-addressing}. 
Recently, sequence-to-sequence LM pre-training has allowed to fully fine-tune QG models (both encoder and decoder), achieving SotA performance \cite{UNILM,qi-etal-2020-prophetnet,UNILMv2,ERNIE-GEN}. Following the latest research in the literature, we focus on sequence-to-sequence LM-based QG models.

QG can be applied to domain adaptation \cite{shakeri-etal-2020-end}, knowledge-enhanced LM pre-training \cite{jia2021question}, adversarial/counterfactual data augmentation \cite{bartolo-etal-2021-improving,paranjape2021retrieval}, and nearest neighbour QA systems \cite{lewis-etal-2021-paq}. Applications of QG go beyond QA, including semantic role labeling \cite{pyatkin-etal-2021-asking}, visual QA \cite{krishna2019information}, multi-hop question decomposition \cite{perez-etal-2020-unsupervised}, and question rewriting \cite{lee-etal-2020-generating}.
Moreover, QG can be applied to unsupervised QA, which consists of training a QA model without any supervision and relying on a QG model to generate questions  \cite{lewis-etal-2019-unsupervised}.
\citet{puri-etal-2020-training} showed that with a carefully-designed QG model, we can generate high-quality QA datasets on which a QA model can even outperform their supervised counterparts.
This inspired \citet{zhang-bansal-2019-addressing} to propose QA-based evaluation, which connects the quality of a QG model to the accuracy of a QA model trained on the synthetic data generated by the QG model.

While QG models can be applied to this variety of tasks, the comparison across tasks is not always straightforward. For this reason, and given the relevance of QG in current research, in this paper we propose an intrinsic QG benchmark in which we can evaluate different aspects of a QG model in a simple manner, including, but not only, analysis of input types, domain adaptability and multilinguality. The most similar work to ours is the MTG benchmark \cite{chen2021mtg}, which contains multilingual test sets for four NLG tasks.
While QG is part of this benchmark, there are a few major differences from our proposed QG-Bench: (i) we provide training/validation/test sets to allow model training in each language in addition to the evaluation; (ii) MTG's test set consists of parallel sentences across languages by a translation from English, while we leverage monolingual datasets; (iii) we include eight languages, while MTG has five; and (iv) QG-Bench includes datasets from different domains and styles.


\section{QG-Bench: A Unified Question Generation Benchmark}
\label{sec:benchmark}

In this section, we describe our process to construct QG-Bench, including data collection and unification (\autoref{sec:collection}), and its statistics (\autoref{sec:dataset-statistics}).




\subsection{Data Collection and Unification}
\label{sec:collection}

We unified a collection of datasets
, designed to be used for QG model training and evaluation.
All datasets are in the same format, where each entry contains four features: \emph{paragraph}, \emph{sentence}, \emph{question}, and \emph{answer}. 
As described in \autoref{fig:task_overview}, we assume \emph{question} as the output of a QG system, which is conditioned by an \emph{answer} and it is always a sub-string of a \emph{sentence} from a \emph{paragraph}.
We leverage existing QA datasets by compiling them into this unified QG format.
All datasets included in QG-Bench are described below.

\noindent \textbf{SQuAD (English).} We first consider SQuAD v1.1 \cite{rajpurkar-etal-2016-squad}, an extractive QA dataset based on Wikipedia which has been used in QG commonly since \cite{du-etal-2017-learning,zhou2017neural}. As the original test set of SQuAD is not released, we use the same data split as in \citep{du-etal-2017-learning}. 

\noindent \textbf{Domain-specific Datasets (English).} To assess models' domain adaptivity, we consider two domain-specific QA datasets: SQuADShifts \cite{miller2020effect} and SubjQA \cite{bjerva-etal-2020-subjqa}.
SQuADShifts contains questions in the same style of SQuAD but from four additional domains (Amazon/Wikipedia/News/Reddit), while SubjQA consists, unlike SQuAD, of subjective questions/answers in general (e.g. \textit{how is the hero?} - \textit{the hero was wonderful}) across six domains. As the original SQuADShifts consists of test set only, we created a new training/validation/test split, in which half of the dataset remains in the test set, while the remaining half is split for validation and training by a 1:2 ratio.

\noindent \textbf{Datasets in Languages other than English.} To investigate multilinguality in QG, we compile the following seven SQuAD-style QA datasets: JAQuAD \cite{so2022jaquad} (Japanese), GerQuAD \cite{GermanQuAD} (German), SberQuAd \cite{efimov2020sberquad} (Russian), KorQuAD \cite{lim2019korquad1} (Korean), FQuAD \cite{dhoffschmidt-etal-2020-fquad} (French), Spanish SQuAD \cite{2016arXiv160605250R} (Spanish), and Italian SQuAD \cite{squad_it} (Italian).
Since they do not release test sets, we sampled a subset from the training sets as the test set following \newcite{du-etal-2017-learning}.
The test sets contain the same number of questions as its validation set, and the new training/test splits have no overlap in terms of the paragraphs.

\noindent \textbf{Other Datasets not Included in QG-Bench.}
In theory, any extractive QA dataset could be part of our benchmark. However, we decided not to include datasets such as BioASQ \cite{tsatsaronis2015overview} and NewsQA \cite{trischler-etal-2017-newsqa} because they have very long input texts, representing another category that needs extra mechanisms to handle long sequences \cite{izacard2020distilling,izacard2020leveraging}, which is out of the scope of this paper. 
In addition, one could leverage multilingual QA benchmarks \cite{clark-etal-2020-tydi,artetxe-etal-2020-cross,lewis-etal-2020-mlqa} to obtain multilingual QG datasets, but XQuAD \cite{artetxe-etal-2020-cross} and MLQA \cite{lewis-etal-2020-mlqa} do not contain training sets, and TydiQA \cite{clark-etal-2020-tydi} contains a very small training set.
Instead, we focused on monolingual QA datasets in each language.

\subsection{Data Statistics}
\label{sec:dataset-statistics}
\autoref{tab:data} summarizes statistics of each QG dataset after unification. It can be observed that SubjQA and SQuADShifts have ten to a hundred times less training data than SQuAD. Also, SubjQA's answers are twice longer than SQuAD's answers, which can be explained by how they differ in the way questions are formed (i.e., SubjQA being more subjective in nature). Likewise, except for Spanish, the datasets for languages other than English contain less training data than the original SQuAD, with the number varying depending on the language. 

\begin{table}[t!]
\centering
\scalebox{0.7}{
\begin{tabular}{@{}l@{\hspace{3pt}}c@{\hspace{3pt}}c@{}}
\toprule
& Data size             & Average character length  \\
& (train/valid/test)    & (para./sent./ques./ans.) \\ \midrule
SQuAD               & 75,722 / 10,570 / 11,877  & 757 / 179 / 59/ 20        \\\midrule
\multicolumn{3}{@{}l}{SubjQA}  \\
- \textit{Book}     & 637 / 92 / 191            & 1,514 / 146 / 28 / 83 \\
- \textit{Elec.}    & 697 / 99 / 238            & 1,282 / 129 / 26 / 66 \\
- \textit{Grocery}  & 687 / 101 / 379           & 896 / 107 / 25 / 49   \\
- \textit{Movie}    & 724 / 101 / 154           & 1,746 / 146 / 27 / 72 \\
- \textit{Rest.}    & 823 / 129 / 136           & 1,006 / 104 / 26 / 51 \\
- \textit{Trip}     & 875 / 143 / 397           & 1,002 / 108 / 27 / 51 \\
\midrule
\multicolumn{3}{@{}l}{SQuADShifts}   \\
- \textit{Amazon}   & 3,295 / 1,648 / 4,942 & 773 / 111 /43 / 18    \\
- \textit{Wiki}     & 2,646 / 1,323 / 3,969 & 773 / 184 / 58 / 26   \\
- \textit{News}     & 3,355 / 1,678 / 5,032 & 781 / 169 / 51 / 20   \\
- \textit{Reddit}   & 3,268 / 1,634 / 4,901 & 774 / 116 / 45 / 19   \\
\midrule
\multicolumn{3}{@{}l}{Multilingual QG} \\
- \textit{Ja}   & 27,809 / 3,939 / 3,939   & 424 / 72 / 32 / 6      \\
- \textit{Es}   & 77,025 / 10,570 / 10,570 & 781 / 122 / 64 / 21    \\
- \textit{De}   & 9,314 / 2,204 / 2,204    & 1,577 / 165 / 59 / 66  \\
- \textit{Ru}   & 40,291 / 5,036 / 5,036   & 754 174 / 64 / 26      \\
- \textit{Ko}   & 54,556 / 5,766 / 5,766   & 521 / 81 / 34 / 6      \\
- \textit{It}   & 46,550 / 7,609 / 7,609   & 807 / 124 / 66 / 16    \\
- \textit{Fr}   & 17,543 / 3,188 / 3,188   & 797 / 160 / 57 / 23    \\
\bottomrule
\end{tabular}
}
\caption{Statistics of of all datasets integrating into our question generation benchmark after unification. }
\label{tab:data}
\end{table}

\section{LMs for Question Generation}
\label{sec:qg-models}

In this section, we formalize the QG task from a language modelling perspective (\autoref{sec:formulation}), including details on the fine-tuning process (\autoref{sec:finetuning-qg}) and the setup for our experiments with QG-Bench (\autoref{sec:para-opt}).

\subsection{Task Formulation}
\label{sec:formulation}

Given an input text $c$, the goal of QG is to generate a natural question $\hat{q}$ related to the information in the input. The task is formulated as a conditional sequence generation, and the model is optimized to maximize the conditional log-likelihood $P(q|c)$ as in \autoref{eq:seq2seq}.
\begin{equation}
    \hat{q} = \argmax_{q} P(q|c)
    \label{eq:seq2seq}
\end{equation}
In practice, the log-likelihood is factorized into word or subword level predictions, similar to other sequence-to-sequence learning settings \cite{sutskever2014sequence}. 

\subsection{Language Model Fine-tuning} 
\label{sec:finetuning-qg}

Fine-tuning sequence-to-sequence LMs on QG can be done in the same way as for Machine Translation or Summarization, where models are trained to predict the output tokens given the input tokens \cite{UNILM,qi-etal-2020-prophetnet,UNILMv2,ERNIE-GEN}.
We follow \citet{chan-fan-2019-recurrent} by introducing a highlight token $\texttt{<hl>}$ to take into account an answer $a$ within a context $c$ as below:
\begin{equation*}
x = [ c_1, \dots, \texttt{<hl>}, a_1, \dots, a_{|a|}, \texttt{<hl>}, \dots, c_{|c|}  ]
\end{equation*}
Instead of a paragraph, we can similarly use a sentence to highlight an answer (sentence-level QG) or highlight a sentence instead of an answer (answer-free QG).
We investigate these model variations in our analysis (\autoref{sec:anlysis_model_input}), but assume the answer highlighted paragraph as the default input.

Note that it is possible to train other types of LMs on QG, but masked LMs were not designed for natural language generation and require a specific decoding technique \cite{chan-fan-2019-recurrent}. Also, recurrent LMs have poor ability for conditional generation on the answer due to its unidirectional architecture \cite{lopez2020transformer}. Since they are not as suited for QG as the sequence-to-sequence models, they are out of the scope of this paper.

\subsection{Experimental Setup}
\label{sec:para-opt}

\noindent \textbf{Comparison Models.} As sequence-to-sequence LMs, we use T5 \cite{T5} and BART \cite{lewis-etal-2020-bart} for the English datasets and mT5 \cite{xue-etal-2021-mt5} and mBART \cite{liu-etal-2020-multilingual-denoising} for the multilingual experiments. 
Model weights are taken from HuggingFace \cite{wolf-etal-2020-transformers}.\footnote{
We use \texttt{t5-small}, \texttt{t5-base}, \texttt{t5-large}, \texttt{facebook/bart-base}, \texttt{facebook/bart-large}, and \texttt{google/mt5-small}.}
Previous research reported improvements on QG with more recent LMs \cite{qi-etal-2020-prophetnet,ERNIE-GEN,UNILMv2}. We tried to replicate these previous works in QG-Bench, but after multiple attempts using their provided code and contacting the authors, this was not possible.  Nonetheless, both T5 and BART are widely used in practice and, as we will show, they can still provide strong results with an appropriate configuration. 

\noindent \textbf{Parameter Optimization.} We performed an extensive exploration to find the best combination of hyper-parameters to fine-tune LMs on QG, which consists of a two-phase search. 
First, we fine-tune a model on every possible configuration from the search space for 2 epochs. The top-5 models in terms of BLEU4 \cite{papineni-etal-2002-bleu} on the validation set are selected to continue fine-tuning until their performance plateaus.\footnote{This two-stage process is introduced due to computation limitations, and we might see further improvements (even if small) if a full validation search is performed.} 
Finally, the model that achieves the highest BLEU4 on the validation set is employed as the final model.
We used BLEU4 as an objective metric in our parameter optimization since it is light to compute, and following previous work \cite{du-cardie-2018-harvesting,UNILM,ERNIE-GEN}. However, as we will see in our experiments, future work could also explore the usage of alternative metrics for validation.
The search space contains 24 configurations, which are made up of learning rates from $[0.0001, 0.00005, 0.00001]$, label smoothing from $[0.0, 0.15]$, and batch size from $[64, 128, 256, 512]$.\footnote{Other parameters are fixed: random seed is $1$, beam size is $4$, input token length is $512$, and output token length is $34$ for fine-tuning and $64$ for evaluation.} Our experiments show that this simple parameter optimization strategy significantly improves all models' performances by robustly finding the best configuration for each one.\footnote{See Appendix for the actual parameters found by the optimization procedure as well as more training details.}

We ran the parameter optimization on a machine equipped with two Nvidia Quadro RTX 8000. Taking SQuAD as a reference, training and evaluation took around three weeks for T5\textsubscript{LARGE}, one week for T5\textsubscript{BASE} and mT5\textsubscript{SMALL}, three days for T5\textsubscript{SMALL}, one week for BART\textsubscript{LARGE}, and four days for BART\textsubscript{SMALL}.


\section{Automatic Evaluation}
\label{sec:autoeval}

In this section, we report the main results in QG-Bench (\autoref{sec:benchmark}), using the methodology described in \autoref{sec:qg-models}.

\begin{table}[!t]
\centering
\scalebox{0.75}{
\begin{tabular}{@{}l@{}r@{\hspace{8pt}}r@{\hspace{8pt}}r@{\hspace{8pt}}r@{\hspace{8pt}}r@{\hspace{8pt}}r@{}}
\toprule
Model                           & Param  & B4 & R-L & MTR & BS & MS \\
\midrule 
NQG (\citeauthor{du-etal-2017-learning})                            & 30M    & 12.28 & 39.75 & 16.62 & -& -  \\
UniLM (\citeauthor{UNILM})     & 340M   & 22.78 & 51.57 & 25.49 & -& -  \\
UniLMv2 (\citeauthor{UNILMv2})                        & 110M   & 24.70 & 52.13 & 26.33 & -& -  \\
ProphetNet (\citeauthor{qi-etal-2020-prophetnet})                     & 340M   & 23.91 & 52.26 & 26.60 & -& -  \\
ERNIE-G (\citeauthor{ERNIE-GEN})  & 340M   & 25.40 & 52.84 & 26.92 & -& - \\
\midrule 
BART\textsubscript{BASE}        & 140M   & 24.68  & 52.66   & 26.05  & 90.87 & 64.47\\
BART\textsubscript{LARGE}       & 400M   & 26.17  & 53.85   & 27.07  & \textbf{91.00}& 64.99\\
T5\textsubscript{SMALL}         & 60M    & 24.40  & 51.43   & 25.84  & 90.45& 63.89\\
T5\textsubscript{BASE}          & 220M   & 26.13  & 53.33   & 26.97  & 90.84& 64.74\\
T5\textsubscript{LARGE}         & 770M   & \textbf{27.21}  & \textbf{54.13}   & \textbf{27.70} & \textbf{91.00}& \textbf{65.29} \\
\bottomrule
\end{tabular}
}
\caption{
QG model fine-tuning results on the test set of SQuAD where the best result in each metric is in bold face.
The results in the top row group are existing SotA models taken from their original papers, while the bottom row contains our models.
}
\label{tab:main-squad}
\end{table}


\subsection{Evaluation Metrics}
To evaluate QG models, BLEU4 \cite[B4,][]{papineni-etal-2002-bleu}, METEOR \cite[MTR,][]{denkowski-lavie-2014-meteor}, and ROUGE\textsubscript{L} \cite[R-L,][]{lin-2004-rouge} are commonly used to compare the generated outputs against reference questions at sentence level. 
We also compute BERTScore \cite[BS,][]{zhang2019bertscore} and MoverScore \cite[MS,][]{zhao-etal-2019-moverscore}. Both leverage BERT-like models on their computation, achieving higher correlations with human judgements than other traditional metrics in various NLG tasks \cite{zhang2019bertscore,zhao-etal-2019-moverscore}. To the best of our knowledge, they have not been applied in QG evaluation before, regardless of their success in NLG. We use the default configuration for both metrics, which make use of RoBERTa\textsubscript{LARGE} \cite{RoBERTa} for BERTScore and DistilBERT\textsubscript{BASE} \cite{DistilBERT} for MoverScore.


\subsection{Results}
\label{sec:main_results}

\noindent \textbf{SQuAD.} \autoref{tab:main-squad} shows our results on the SQuAD test set along with other reported results from the literature. T5\textsubscript{LARGE} provides the best results overall according to all automatic metrics. Even parameter-efficient models such as T5\textsubscript{BASE} outperform ERNIE-GEN \cite{ERNIE-GEN}, and T5\textsubscript{SMALL} performs competitively with UniLMv2 \cite{UNILMv2} with nearly half the parameters. UniLMv2, in particular, was proposed as a highly-effective model in spite of its light weight. According to these results, T5\textsubscript{SMALL} is also competitive on the QG task while being significantly lighter than other models.
While T5 attains the best overall results, BART also proves competitive. In fact, BART\textsubscript{BASE} is slightly better than T5\textsubscript{BASE} and BART\textsubscript{LARGE} is equal to T5\textsubscript{LARGE} according to BERTScore. 
In general, it is hard to reliably compare different model architectures for the QG tasks, as there are different possible confounding factors including the model size. To have a more complete picture on the final performance, we complement this initial automatic evaluation in SQuAD with an extensive manual evaluation in \autoref{sec:result_manual_evaluation}.
\begin{table}[t!]
\centering
\scalebox{0.75}{
\begin{tabular}{llrrrrr}
\toprule
    & Model & B4    & R-L   & MTR   & BS    & MS    \\\midrule
\multirow{3}{*}{\rotatebox{90}{English}}
& mT5\textsubscript{SMALL}  & 21.65 & 48.95 & 23.83 & 90.01 & 62.75 \\
& mT5\textsubscript{BASE}   & \textbf{23.03} & \textbf{50.67} & \textbf{25.18} & 90.23 & 63.60 \\
& mBART                     & \textbf{23.03} & 50.58 & 25.10 & \textbf{90.36} & \textbf{63.63} \\\midrule
\multirow{3}{*}{\rotatebox{90}{Russian}}
& mT5\textsubscript{SMALL}  & 16.31 & 31.39 & 26.39 & 84.27 & 62.49 \\
& mT5\textsubscript{BASE}   & 17.63 & 33.02 & 28.48 & 85.82 & 64.56 \\
& mBART                     & \textbf{18.80} & \textbf{34.18} & \textbf{29.30} & \textbf{87.18} & \textbf{65.88} \\\midrule
\multirow{3}{*}{\rotatebox{90}{Japanese}}
& mT5\textsubscript{SMALL}  & 30.49 & 50.88 & 29.03 & 80.87 & 58.67 \\
& mT5\textsubscript{BASE}   & \textbf{32.54} & 52.67 & \textbf{30.58} & 81.77 & 59.68 \\
& mBART                     & 32.16 & \textbf{52.95} & 29.97 & \textbf{82.26} & \textbf{59.88} \\\midrule
\multirow{3}{*}{\rotatebox{90}{Italian}}
& mT5\textsubscript{SMALL}  & 7.37  & 21.93 & 17.57 & 80.80 & 56.79 \\
& mT5\textsubscript{BASE}   & \textbf{7.70}  & \textbf{22.51} & \textbf{18.00} & \textbf{81.16} & \textbf{57.11} \\
& mBART                     & 7.13  & 21.69 & 17.97 & 80.63 & 56.84 \\\midrule
\multirow{3}{*}{\rotatebox{90}{Korean}}
& mT5\textsubscript{SMALL}  & 10.57 & 25.64 & 27.52 & 82.89 & 82.49 \\
& mT5\textsubscript{BASE}   & \textbf{12.18} & \textbf{28.57} & 29.62 & \textbf{84.52} & \textbf{83.36} \\
& mBART                     & 10.92 & 27.76 & \textbf{30.23} & 83.89 & 82.95 \\\midrule
\multirow{3}{*}{\rotatebox{90}{Spanish}}
& mT5\textsubscript{SMALL}  & 9.61  & 24.62 & 22.71 & 84.07 & 59.06 \\
& mT5\textsubscript{BASE}   & \textbf{10.15} & \textbf{25.45} & \textbf{23.43} & \textbf{84.47} & \textbf{59.62} \\
& mBART                     & 9.18  & 24.26 & 22.95 & 83.58 & 58.91 \\\midrule
\multirow{3}{*}{\rotatebox{90}{German}}
& mT5\textsubscript{SMALL}  & 0.43  & 10.08 & 11.47 & 79.90 & 54.64 \\
& mT5\textsubscript{BASE}   & \textbf{0.87}  & 11.10 & 13.65 & 80.39 & 55.73 \\
& mBART                     & 0.75  & \textbf{11.19} & \textbf{13.71} & \textbf{80.77} & \textbf{55.88} \\\midrule
\multirow{3}{*}{\rotatebox{90}{French}}
& mT5\textsubscript{SMALL}  & \textbf{8.55}  & \textbf{28.56} & \textbf{17.51} & \textbf{80.71} & \textbf{56.50} \\
& mT5\textsubscript{BASE}   & 6.14  & 25.88 & 15.55 & 77.81 & 54.58 \\
& mBART                     & 0.72  & 16.40 & 7.78  & 71.48 & 50.35 \\\bottomrule
\end{tabular}
}
\caption{
QG model fine-tuning results on the test set of all language-specific QG-Bench datasets where the best result in each language is in bold face.
}
\label{tab:main-mlqg}
\end{table}

\noindent \textbf{Language-specific Datasets.} \autoref{tab:main-mlqg} presents the results on each language-specific dataset in QG-Bench with mT5\textsubscript{SMALL}, mT5\textsubscript{BASE}, and mBART. As this work introduces the first ever comprehensive multilingual QG model training/evaluation, these results can be viewed as baselines for future work in multilingual QG.
Compared to results in English SQuAD, scores in multilingual QG are mostly worse than the smallest English model (T5\textsubscript{SMALL}), which showcases the difficulties of non-English QG.
Some languages are notably underperforming, which can be partially explained by the size of the training set.
As we see in \autoref{sec:dataset-statistics}, some datasets such as German and French have a limited amount of training instances, resulting in underfitting models for those languages.
In general, the low scores in non-English datasets can be attributed to the underlying model, so scaling up the model could lead to better performances in future work.

\begin{table}[!t]
\centering
\scalebox{0.75}{
\begin{tabular}{@{}l@{\hspace{5pt}}@{}l@{\hspace{5pt}}@{}l@{\hspace{5pt}}@{}r@{\hspace{5pt}}@{\hspace{5pt}}r@{\hspace{5pt}}@{\hspace{5pt}}r@{\hspace{5pt}}@{\hspace{5pt}}r@{}@{\hspace{5pt}}@{\hspace{5pt}}r@{}}
\toprule
\multicolumn{2}{@{}l}{Domain} & Model & B4 & R-L & MTR & BS & MS \\\midrule
\multirow{20}{*}{\rotatebox{90}{SQuADShifts}} 
                            & \multirow{5}{*}{Amazon}   & BART\textsubscript{BASE}  & 9.92          & 27.94         & 22.78         & \textbf{92.77} & 63.25\\
                            &                           & BART\textsubscript{LARGE} & 9.80          & 28.69         & 23.79         & 92.49 & 63.31\\
                            &                           & T5\textsubscript{SMALL}   & 8.41          & 27.04         & 22.17         & 91.89 & 62.11 \\
                            &                           & T5\textsubscript{BASE}    & 9.80          & 28.94         & 23.85         & 92.43 & 63.27 \\
                            &                           & T5\textsubscript{LARGE}   & \textbf{10.42}& \textbf{29.51}& \textbf{24.39}& 92.65 & \textbf{63.71} \\
                            \cline{2-8}
                            & \multirow{5}{*}{Wiki}     & BART\textsubscript{BASE}  & 11.50 & 29.00 & 26.60 & 93.12 & 65.86\\
                            &                           & BART\textsubscript{LARGE} & \textbf{12.12} & 29.94 & 27.12 & \textbf{93.39} & \textbf{66.22}\\
                            &                           & T5\textsubscript{SMALL}   & 10.90 & 28.18 & 25.95 & 92.63 & 65.04 \\
                            &                           & T5\textsubscript{BASE}    & 11.67 & 29.49 & 27.04 & 93.07 & 65.94 \\
                            &                           & T5\textsubscript{LARGE}   & 12.04 & \textbf{30.10} & \textbf{27.67} & 93.13 & 66.31 \\
                            \cline{2-8}
                            & \multirow{5}{*}{News}     & BART\textsubscript{BASE}  & 8.78 & 24.85 & 25.13 & 92.86 & 64.99\\
                            &                           & BART\textsubscript{LARGE} & 8.74 & 25.28 & 25.08 & \textbf{93.04} & 65.02\\
                            &                           & T5\textsubscript{SMALL}   & 7.71  & 23.43 & 23.70 & 92.20 & 63.71 \\
                            &                           & T5\textsubscript{BASE}    & 8.53  & 24.93 & 25.21 & 92.68 & 64.70 \\
                            &                           & T5\textsubscript{LARGE}   & \textbf{9.16}  & \textbf{25.97} & \textbf{25.98} & 93.01 & \textbf{65.46} \\
                            \cline{2-8}
                            & \multirow{5}{*}{Reddit}   & BART\textsubscript{BASE}  & 8.78 & 26.03 & 22.57 & 92.32 & 62.35\\
                            &                           & BART\textsubscript{LARGE} & \textbf{9.31} & \textbf{27.31} & 23.75 & 92.50 & 62.64\\
                            &                           & T5\textsubscript{SMALL}   & 7.60  & 24.90 & 21.90 & 91.70 & 61.39 \\
                            &                           & T5\textsubscript{BASE}    & 8.75  & 26.84 & 23.57 & 92.26 & 62.52 \\
                            &                           & T5\textsubscript{LARGE}   & 9.16  & 27.24 & \textbf{23.97} & \textbf{92.43} & \textbf{62.74} \\
                            \midrule
\multirow{30}{*}{\rotatebox{90}{SubjQA}}      
                            & \multirow{5}{*}{Book}     & BART\textsubscript{BASE}  & \textbf{2.03} & 23.24 & 20.57 & 92.96 & 62.85\\
                            &                           & BART\textsubscript{LARGE} & 0.00  & \textbf{23.71} & 20.6 & 92.84 & 62.45\\
                            &                           & T5\textsubscript{SMALL}   & 0.00  & 19.77 & 18.52 & 92.40 & 61.46 \\
                            &                           & T5\textsubscript{BASE}    & 0.00  & 22.95 & \textbf{21.20} & \textbf{93.32} & \textbf{63.14} \\
                            &                           & T5\textsubscript{LARGE}   & 0.00  & 23.68 & 20.83 & 92.89 & 62.51 \\
                            \cline{2-8}
                            & \multirow{5}{*}{Elec.}    & BART\textsubscript{BASE}  & 3.83 & 29.41 & 25.08 & 93.76 & 66.00\\
                            &                           & BART\textsubscript{LARGE} & \textbf{5.18} & 28.87 & 25.17 & 93.51 & 65.68\\
                            &                           & T5\textsubscript{SMALL}   & 0.00  & 29.65 & 26.95 & 94.18 & 68.29 \\
                            &                           & T5\textsubscript{BASE}    & 4.55  & 29.99 & 27.39 & 94.26 & 68.33 \\
                            &                           & T5\textsubscript{LARGE}   & 4.57  & \textbf{30.55} & \textbf{27.56} & \textbf{94.27} & \textbf{68.80} \\
                            \cline{2-8}
                            & \multirow{5}{*}{Grocery}  & BART\textsubscript{BASE}  & 1.82 & \textbf{24.54} & 20.8 & \textbf{94.09} & 65.76\\
                            &                           & BART\textsubscript{LARGE} & \textbf{1.93} & 24.28 & 20.42 & 94.1 & \textbf{65.79}\\
                            &                           & T5\textsubscript{SMALL}   & 0.00  & 22.17 & \textbf{23.31} & 93.24 & 65.64 \\
                            &                           & T5\textsubscript{BASE}    & 0.83  & 15.63 & 19.87 & 90.56 & 61.47 \\
                            &                           & T5\textsubscript{LARGE}   & 1.13  & 17.40 & 20.64 & 91.39 & 63.41 \\
                            \cline{2-8}
                            & \multirow{5}{*}{Movie}    & BART\textsubscript{BASE}  & 3.89 & 25.43 & 20.55 & 93.61 & 62.91\\
                            &                           & BART\textsubscript{LARGE} & \textbf{4.21} & 25.92 & 21.64 & 93.23 & 62.4\\
                            &                           & T5\textsubscript{SMALL}   & 0.00  & 25.76 & 22.54 & 94.08 & 64.63 \\
                            &                           & T5\textsubscript{BASE}    & 2.65  & \textbf{26.33} & \textbf{23.11} & \textbf{94.13} & \textbf{64.91} \\
                            &                           & T5\textsubscript{LARGE}   & 0.00  & 25.06 & 21.70 & 93.64 & 63.88 \\
                            \cline{2-8}
                            & \multirow{5}{*}{Rest.}    & BART\textsubscript{BASE}  & 3.43 & 24.26          & 21.35 & \textbf{93.23} & 62.67\\
                            &                           & BART\textsubscript{LARGE} & \textbf{5.54} & 24.77          & \textbf{22.46} & 93.23 & \textbf{63.57}\\
                            &                           & T5\textsubscript{SMALL}   & 0.00  & 11.72          & 13.21 & 87.81 & 55.42 \\
                            &                           & T5\textsubscript{BASE}    & 0.00  & 11.96         & 14.75 & 88.48 & 56.19 \\
                            &                           & T5\textsubscript{LARGE}   & 4.19  & \textbf{24.94} & 21.99 & 93.22 & 63.25 \\
                            \cline{2-8}
                            & \multirow{5}{*}{Trip}     & BART\textsubscript{BASE}  & 4.79 & 26.37 & 25.26 & 93.92 & 64.91\\
                            &                           & BART\textsubscript{LARGE} & \textbf{5.66} & 26.5 & 24.32 & 93.85 & 64.02\\
                            &                           & T5\textsubscript{SMALL}   & 2.49  & 23.91 & 25.56 & 93.75 & 66.57 \\
                            &                           & T5\textsubscript{BASE}    & 1.74  & 16.06 & 20.13 & 90.76 & 59.70 \\
                            &                           & T5\textsubscript{LARGE}   & 5.35  & \textbf{27.69} & \textbf{27.45} & \textbf{94.46} & \textbf{67.76}  \\
                            \bottomrule
\end{tabular}
}
\caption{
QG model fine-tuning results on the test set of SQuADShifts and SubjQA where the best result in each metric is in bold face.
}
\label{tab:main-subjqa-squadshifts}
\end{table}
\noindent \textbf{Domain-specific Datasets.} \autoref{tab:main-subjqa-squadshifts} shows the results from all domain-specific datasets included in QG-Bench: SQuADShifts and SubjQA.
Since each domain contains a small training set, our main strategy to achieve domain-specific QG models is to initialize their weights with a SQuAD fine-tuned model, and continue fine-tuning on the domain-specific training set (more details on different strategies in \autoref{sec:domainadapt}).
As expected, given the subjective nature of the dataset, results on SubjQA are generally low for most metrics, except for BERTScore whose score is even higher than in SQuAD in some cases. This implies that a model's prediction may have less word-overlap against the true question, while its semantics is close to the true question to some extent.

\section{Analysis}

In this section, we complement the automatic evaluation with an extensive analysis on various relevant aspects of the question generation models.

\subsection{Model Input}
\label{sec:anlysis_model_input}

\begin{figure}[t]
 \centering
 \includegraphics[width=0.8\columnwidth]{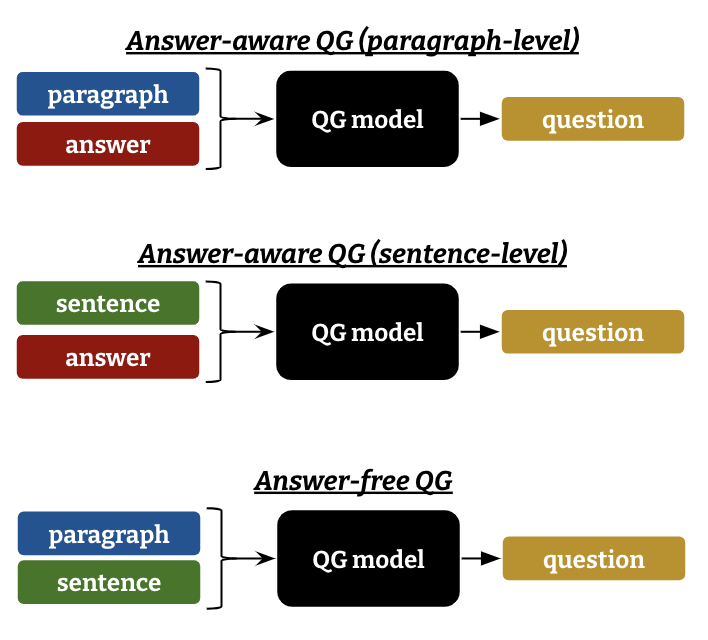}
\caption{Input variations of QG models.}
 \label{fig:model_type}
\end{figure}

In our main experiments, the model input is the paragraph in which the answer is highlighted, as described in \autoref{sec:finetuning-qg}.
Here we explore variations of the QG model's input type to understand the effect of different types of context.
Concretely we consider two additional variants: \emph{sentence-level} models that only take as input the sentence that contains the answer (instead of the whole paragraph); and \emph{answer-free} models that highlight the sentence in the paragraph instead of the answer. \autoref{fig:model_type} provides a summary of the three different input types analysed.

In \autoref{tab:input_variation} we report automatic metrics from answer-free models and sentence-level QG models on SQuAD. In general, paragraph-based models, which use the most complete input, attain the best overall results.
For example, answer-free T5\textsubscript{LARGE} performs worse than paragraph-level T5\textsubscript{SMALL} in all the metrics except METEOR, which indicates the importance of the answer at question generation. Nonetheless, not having the answer as input provides competitive results, which may appear to be surprising given the incomplete input. When comparing sentence-level and paragraph-level, the difference is reduced, but paragraph-level models consistently outperform their sentence-level counterparts, even when smaller models are used. 
This implies that models actually utilize the global context provided by the full paragraph when it is available, rather than the more local information within the sentence only.

\begin{table}[t]
\centering
\scalebox{0.75}{
\begin{tabular}{@{}l@{\hspace{4pt}}l@{\hspace{8pt}}r@{\hspace{8pt}}r@{\hspace{8pt}}r@{\hspace{8pt}}r@{\hspace{8pt}}r@{}}
\toprule
\multicolumn{2}{@{}l}{Model} & B4 & R-L & MTR & BS & MS \\\midrule
\multirow{5}{*}{\rotatebox{90}{{Answer-free}}}
& BART\textsubscript{BASE}  & 21.97               & 49.70                 & 23.72       & 90.38 & 63.07 \\
& BART\textsubscript{LARGE} & 23.47               & 50.25                 & 24.94       & 90.28 & 63.28 \\
& T5\textsubscript{SMALL}   & 21.12               & 47.47                 & 23.38       & 89.64 & 62.07 \\
& T5\textsubscript{BASE}    & 22.86               & 49.51                 & 24.52       & 90.03 & 62.99 \\
& T5\textsubscript{LARGE}   & 24.27               & 51.30                 & 25.67       & 90.41 & 63.97 \\\midrule
\multirow{5}{*}{\rotatebox{90}{{Sent-level}}}
& BART\textsubscript{BASE}  & 23.86               & 51.43                 & 25.18       & 90.70 & 63.85 \\
& BART\textsubscript{LARGE} & 23.86               & 51.43                 & 25.18       & 90.70 & 63.85 \\
& T5\textsubscript{SMALL}   & 23.23               & 50.18                 & 24.80       & 90.36 & 63.18 \\
& T5\textsubscript{BASE}    & 24.33               & 51.81                 & 25.81       & 90.73 & 64.00 \\
& T5\textsubscript{LARGE}   & 25.36               & 52.53                 & 26.28       & 90.88 & 64.44 \\\midrule
\multirow{5}{*}{\rotatebox{90}{{Para-level}}} 
& BART\textsubscript{BASE}  & 24.68  & 52.66   & 26.05  & 90.87 & 64.47\\
& BART\textsubscript{LARGE} & 26.17  & 53.85   & 27.07  & \textbf{91.00}& 64.99\\
& T5\textsubscript{SMALL}   & 24.40  & 51.43   & 25.84  & 90.45& 63.89\\
& T5\textsubscript{BASE}    & 26.13  & 53.33   & 26.97  & 90.84& 64.74\\
& T5\textsubscript{LARGE}   & \textbf{27.21}  & \textbf{54.13}   & \textbf{27.70} & \textbf{91.00}& \textbf{65.29} \\
\bottomrule
\end{tabular}
}
\caption{
QG model fine-tuning results on the test set of SQuAD for answer-free and sentence/paragraph-level QG models. The best overall result for each metric is in boldface.
}
\label{tab:input_variation}
\end{table}

\subsection{Manual Evaluation}
\label{sec:result_manual_evaluation}

Given the limitations of automatic metrics in text generation research \cite{reiter-2018-structured,bhandari-etal-2020-evaluating,alva-manchego-etal-2021-un}, we also conducted a manual evaluation using Amazon Mechanical Turk, focusing on three criteria: 
\emph{grammaticality} (i.e.\ grammatical correctness), \emph{understandability} (i.e.\ whether the question is easy to be understood by readers) and \emph{answerability} (i.e.\ whether the question can be answered by the given input answer).\footnote{Understandability could correlate with grammaticality, but a question without any grammatical mistakes can have low understandability due to an over complex structure. Likewise, a question can be understandable even with a few grammatical mistakes. Annotation guidelines are included in the Appendix.}
We randomly sampled 500 unique paragraphs from the SQuAD test set and selected a single answer in each paragraph. For each of the 500 paragraph-answer pairs, we generated questions from six QG models, and asked human annotators to score them for the criteria with a 3-points scale. Each question was evaluated by five judges, thus collecting a total of 15,000 human judgments. As quality control, we asked workers to be native English speakers, and instructed them to do a qualification test first, and only those who passed the test worked on our annotation task.
The given time of each assignment (with each assignment containing ten instances to annotate) was 30 minutes, and the reward of the 
annotation task was \$2 per assignment.\footnote{The full price of annotation exercise was about \$3,000.} We attach a screenshot of the annotation interface in the Appendix. 

\noindent \textbf{Comparison Models.} For the manual evaluation, the target QG models include T5\textsubscript{LARGE}, T5\textsubscript{SMALL} and BART\textsubscript{LARGE} based paragraph-level QG models;  T5\textsubscript{LARGE} sentence-level and answer-free QG models; and NQG \cite{du-etal-2017-learning}, which is based on an LSTM-architecture. NQG is included for completeness and to better analyse the effect of pre-trained LMs in general. T5\textsubscript{LARGE} is our best model according to automatic metrics, so we compare it against different input types (answer-free or sentence-level), different sizes (T5\textsubscript{SMALL}), and different model architectures (BART\textsubscript{LARGE}). 

\noindent \textbf{Inter-annotator Agreement.} Since there are five unique annotators per each generated question, we calculated Fleiss's kappa to measure the inter-annotator-agreement. 
We obtained 0.30 and 0.36 for grammaticality and understandability respectively, resulting in fair-agreement \cite{landis1977measurement}. The kappa is 0.61 in answerability, which is a substantial-agreement.

\begin{table}[t]
\centering
\scalebox{0.75}{
\begin{tabular}{@{}l@{\hspace{5pt}}r@{\hspace{5pt}}r@{\hspace{5pt}}r@{\hspace{5pt}}r@{\hspace{5pt}}r@{\hspace{5pt}}r@{\hspace{5pt}}r@{\hspace{5pt}}r@{}}
\toprule
\multirow{2}{*}{Model}  & \multicolumn{3}{c}{Manual Metric} & \multicolumn{5}{c}{Automatic Metric}\\
 & Ans. & Gra. & Und. & B4 & R-L & MTR & BS & MS\\\midrule
NQG                         & 1.21  & 2.35  & 2.63  & 3.33  & 14.30  & 33.53    & 88.27 & 58.25 \\
BART\textsubscript{LARGE}   & 2.70  & 2.89  & 2.93  & 16.15 & 29.93  & 51.35    & \textbf{90.95} & 65.44 \\
T5\textsubscript{SMALL}     & 2.51  & 2.83  & 2.90  & 13.43 & 27.38  & 48.86    & 90.41 & 64.27 \\
T5\textsubscript{LARGE}     & \textbf{2.80}  & \textbf{2.93}  & \textbf{2.95}  & \textbf{17.56} & \textbf{30.42}  & \textbf{52.00}    & 90.94 & \textbf{66.09} \\
 - sent-level               & 2.47  & 2.91  & \textbf{2.95}  & 14.88 & 27.49  & 48.97    & 90.76 & 64.53 \\
 - answer-free              & 2.46  & 2.91  & \textbf{2.95}  & 13.62 & 26.82  & 47.37    & 90.20 & 64.00 \\
\bottomrule
\end{tabular}
}
\caption{Manual evaluation results along with the automatic metrics. Each score is averaged within the 500 questions for the evaluation where the best result in each metric is in bold face.
}
\label{tab:manual_evaluation}
\end{table}

\noindent \textbf{Model-wise Evaluation.}
We report the results of our manual evaluation in \autoref{tab:manual_evaluation}, where each score is averaged over the 500 questions used in the study. 
Answerability is the most affected by model size/context and type/model architecture, compared to the other metrics, except for NQG, which is the only non-LM pre-training based approach. 
In fact, when we compare T5\textsubscript{LARGE}'s paragraph-level against sentence-level, answerability decreases unlike the other two criteria, highlighting the importance of including all relevant context available so that the model can generate a suitable question. 
On the other hand, while answer-free models are worse than sentence-level models according to automatic metrics, the manual evaluation does not reflect a significant difference between them. In general, we can see how T5\textsubscript{LARGE}, which is the best model overall according to the automatic metrics, is also the most robust model overall according to the manual evaluation, which reinforces the conclusions from the automatic evaluation.

\begin{figure}[t]
    \centering
    \includegraphics[width=\columnwidth]{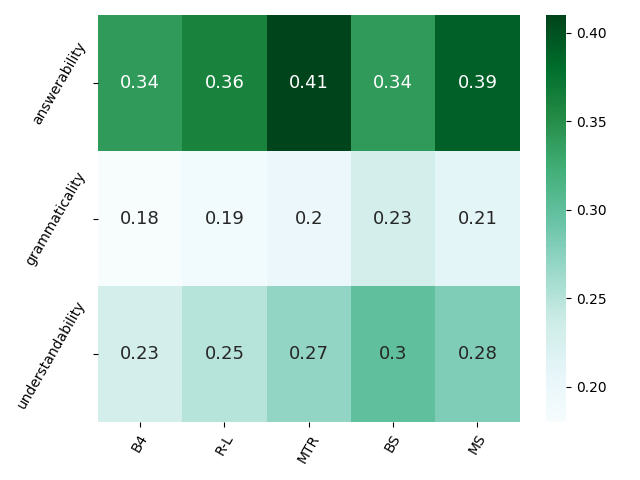}
    \caption{Spearman's rank correlation over all the generated questions within the manual evaluation.}
    \label{fig:corr}
\end{figure}

\noindent \textbf{Correlation Analysis.}
Leveraging the large dataset of collected human judgments, we investigate the correlation between human annotations and the automatic metrics considered in the automatic evaluation (\autoref{sec:main_results}). For this analysis, we included all the generated questions from all the models considered in the manual evaluation. This means 3,000 generated questions from six diverse models where each question receives five annotations. We took the average across all the five annotators for each generated question to compute the correlation. 
\autoref{fig:corr} shows the Spearman's rank correlation coefficient across the automatic metrics and the criteria collected through our manual evaluation.\footnote{See the full correlation analysis in Appendix \autoref{app:correlation}.}  The p-values of all correlations are less than 0.05, so they are all statistically significant.
To check the significance of the increase in the correlation across metrics, we ran a William test, showing that the increase is statistically significant in all cases.\footnote{Full results of the William test are in Appendix \autoref{app:williams}.}

According to the correlation analysis, no metric achieved a high agreement with human judgements in all criteria.
This means that we should not rely on a single metric to capture all quality aspects of a model's output. 
We can conclude, however, that METEOR and MoverScore are well-aligned with human judgements on answerability, while BERTScore appears to be better suited for grammaticality and understandability.
Most importantly, BLEU4 and ROUGE\textsubscript{L}, which have been mostly used in the QG literature as default metrics, are not as reliable as the other metrics in any criteria.

\subsection{Domain Adaptation}
\label{sec:domainadapt}

\begin{figure}[t]
    \centering
    \includegraphics[width=\columnwidth]{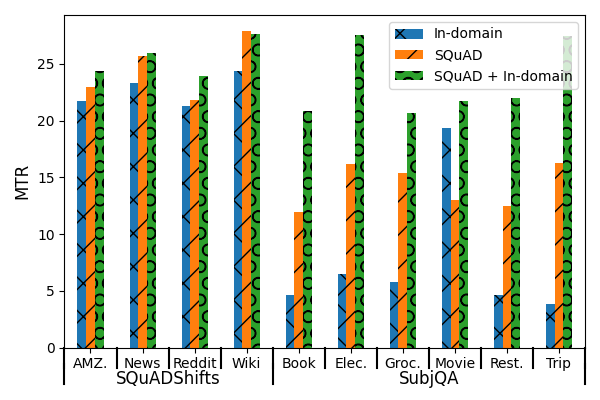}
    \caption{Comparison of METEOR (MTR) scores for T5\textsubscript{LARGE} across in-domain fine-tuning, zero-shot transfer of SQuAD fine-tuned model, and in-domain fine-tuning from SQuAD model.}
    \label{fig:domain-difference}
\end{figure}

In our main experiments in the domain-specific datasets of QG-Bench (\autoref{sec:main_results}), models were initialized by the SQuAD fine-tuned model due to the limited training set in each domain.
To further explore the domain adaptability of QG models, we compared three different setups: (1) fine-tuning in the in-domain training set without SQuAD initialization, (2) zero-shot transfer from the SQuAD fine-tuned model, and (3) fine-tuning with a prior SQuAD initialization.
\autoref{fig:domain-difference} shows the results of T5\textsubscript{LARGE} (the best model in most of the domains in \autoref{tab:main-subjqa-squadshifts} and the manual evaluation) in each domain for those three settings. For this analysis, we focus on the METEOR metric, which attains the highest correlation with human judges in answerability.\footnote{The full set of results for other metrics and models is available in the Appendix, with similar general trends.} We can confirm that the best setup is to initialize the model on SQuAD and then further fine-tune it on the domain-specific training sets. For SQuADShifts, however, this improvement is less marked in general, suggesting that T5 can handle inputs from different domains to a certain extent.
In contrast, the zero-shot setting with SQuAD fine-tuning in SubjQA achieves very poor results overall. This is to a certain extent expected since the questions in SubjQA are of very different styles. 

Finally, while in this section we focused on the domain adaptability for English, in the Appendix we also show zero-shot cross-lingual transfer results, adapting English-training models to other languages.
Similarly to previous work \cite{chen2021mtg}, the main conclusion is that there is still significant room for improvement for zero-shot cross-lingual transfer in QG.

\section{Conclusion}

In this paper we presented QG-Bench, a unified benchmark and evaluation for testing paragrah-level QG models. The benchmark is composed of the general-purpose SQuAD dataset, as well as domain-specific datasets of different styles for English. Moreover, it includes language-specific datasets for eight different languages. Using QG-Bench as a reference, we tested recent generative language models on the task, and evaluated them across a range of automatic metrics. To complement the automatic evaluation, we performed a comprehensive manual evaluation to better understand the performance of models and the role of automatic metrics (e.g., our study shows there are better metrics than the popular BLEU4 when it comes to QG). In general, our results show that LMs have come a long way for QG, being very competitive (e.g., T5 attains an overall manual score of, respectively, 2.80, 2.93 and 2.95 in answerability, grammaticality and understandability on SQuAD), but have room for improvement when dealing with different domains and styles, and especially on languages other than English.

As future work, we will continue to study QG evaluation metrics in-depth to better understand what aspects we are missing when we use specific automatic metrics, using our manual evaluation as a proxy.
Moreover, the QG models analysed in this paper require an answer to be specified beforehand to generate the question. As a way to relax the constrain, we can train models for question and answer pair generation (QAG) by generating the answer together with the question given a context. By generating both answers and questions together, new evaluation metrics would also be required to understand the validity and diversity of the answers selected, which we leave for future work.


\section*{Limitations}
In this paper, we explored paragraph-level QG models, which limits their input up to around 500 tokens, and the same methodology cannot be easily applied to longer documents.
In multilingual QG modeling, we considered datasets in seven different languages, but all of them are medium- to high-resource languages, so our experimental results cannot be generalized to a truly low-resource language setting.
Finally, although the focus of our paper is mostly in SQuAD-style one hop extractive QA, QG is also studied in more complex scenarios such as multi-hop QG with graph neural networks \cite{pan-etal-2020-semantic} and QG for very long answers \cite{cao-wang-2021-controllable}.
Moreover, QG models are used to attain better interpretability in question answering such as multi-hop question decomposition \cite{perez-etal-2020-unsupervised} and question rewriting \cite{lee-etal-2020-generating}.
As future work, we will expand our analysis to more complex scenarios and explore the connectivity with the QA task.

\section*{Ethics Statement}
As the potential risk at using our QG models, it has been reported that language models inherit undesirable biases and generate toxic language \cite{schick-etal-2021-self}, and one could find such text in the generated question. However, we internally checked the generated questions used for the manual evaluation, and confirmed that they did not contain toxic content.

\section*{Acknowledgements}
Jose Camacho-Collados is supported by a UKRI Future Leaders Fellowship.

\bibliography{anthology,custom}
\bibliographystyle{acl_natbib}

\clearpage
\appendix

\section{Parameter Optimization}

\subsection{Best Parameters}
\begin{table}[h]
\centering
\scalebox{0.7}{
\begin{tabular}{@{}l@{\hspace{5pt}}c@{\hspace{5pt}}c@{\hspace{5pt}}c@{\hspace{5pt}}c@{\hspace{5pt}}c@{}}
\toprule
\multirow{2}{*}{Model} & \multicolumn{1}{c}{\multirow{2}{*}{Epoch}} & Learning & \multirow{2}{*}{Batch} & Gradient & Label     \\
                       & \multicolumn{1}{c}{}                       & Rate     &                        & Steps    & Smoothing   \\ \midrule
\multicolumn{6}{@{}l}{\textit{Answer-aware Model (paragraph-level)}} \\
BART\textsubscript{BASE}&  7     & 0.0001   & 32    & 8        & 0.15           \\
BART\textsubscript{LARGE}& 4     & 0.00005 & 32    & 4        & 0.15            \\
T5\textsubscript{SMALL}& 9     & 0.0001   & 64    & 1        & 0.15            \\
T5\textsubscript{BASE}& 5     & 0.0001   & 16    & 4        & 0.15           \\
T5\textsubscript{LARGE}& 6     & 0.00005 & 16    & 4        & 0.15           \\\midrule
\multicolumn{6}{@{}l}{\textit{Answer-aware Model (sentence-level)}} \\
BART\textsubscript{BASE}&  3     & 0.0001   & 64    & 2        & 0.15           \\
BART\textsubscript{LARGE}& 8     & 0.00005 & 32    & 16        & 0.15            \\
T5\textsubscript{SMALL}& 8     & 0.0001   & 64    & 1        & 0.15            \\
T5\textsubscript{BASE}& 8     & 0.0001   & 64    & 1        & 0.15           \\
T5\textsubscript{LARGE}& 6     & 0.00005 & 16    & 4        & 0.15           \\\midrule
\multicolumn{6}{@{}l}{\textit{Answer-free Model}} \\
BART\textsubscript{BASE}&  4     & 0.0001   & 32    & 8        & 0.15           \\
BART\textsubscript{LARGE}& 4     & 0.00005 & 32    & 4        & 0.15            \\
T5\textsubscript{SMALL}& 7     & 0.0001   & 64    & 4        & 0.15            \\
T5\textsubscript{BASE}& 8     & 0.0001   & 16    & 4        & 0.15           \\
T5\textsubscript{LARGE}& 7     & 0.00005 & 16    & 4        & 0.15           \\\bottomrule
\end{tabular}
}
\caption{
The best parameter to fine-tune each model on SQuAD we found through the parameter optimization.
}
\label{tab:parameter}
\end{table}

\autoref{tab:parameter} shows the best configuration to fine-tune each model that we obtain through the parameter optimization process. To fine-tune T5 model, we use the task prefix \texttt{generate question:} at the beginning of the input text.

\subsection{Fine-tuning without Optimization}
\begin{table}[h]
\centering
\scalebox{0.75}{
\begin{tabular}{@{}l@{\hspace{5pt}}r@{\hspace{10pt}}r@{\hspace{10pt}}r@{\hspace{10pt}}r@{\hspace{10pt}}r@{}}
\toprule
Model        & B4 & R-L & MTR & BS & MS \\ \midrule
BART\textsubscript{BASE}  & -0.28                  & -0.17                   & -0.07                   & -0.01                  & 0.00                   \\
BART\textsubscript{LARGE} & -2.22                  & -1.65                   & -1.16                   & -0.06                  & -0.57                  \\
T5\textsubscript{SMALL}   & -1.73                  & -1.89                   & -1.16                   & -0.28                  & -0.82                  \\
T5\textsubscript{BASE}    & -0.72                  & -0.58                   & -0.39                   & -0.10                  & -0.28                  \\
T5\textsubscript{LARGE}   & -0.18                  & -0.15                   & 0.00                    & -0.07                  & -0.09              \\\bottomrule
\end{tabular}
}
\caption{Decrease in automatic metrics of our QG models without parameter optimization .}
\label{tab:default}
\end{table}

\autoref{tab:default} shows the decrease in each metric for SQuAD if the model is fine-tuned without parameter optimization.\footnote{We follow the hyperparameter used to fine-tune ERNIE-GEN on SQuAD QG in the original paper.}
We observe decent drops in performance. T5\textsubscript{SMALL} and BART\textsubscript{LARGE} lose around 2 points in BLEU4 and ROUGE\textsubscript{L}. 
According to these results, we infer that T5 and BART were worse than more recent LMs (ProphetNet, UniLM, or ERNIE-GEN) in QG just because they were under-fitted to the task due to  sub-optimal fine-tuning parameters, rather than they being inferior to those recent LMs in terms of learning the QG task.

\section{Manual Evaluation}
\subsection{Sample Outputs}
\begin{table*}[h]
\centering
\scalebox{0.5}{
\begin{tabular}{@{}p{0.7\linewidth}|p{0.2\linewidth}|l|p{0.3\linewidth}|ccc@{}}
\toprule
Paragraph & Question & Model & Prediction & Answerability & Grammaticality & Understandability \\\midrule
\multirow{6}{\linewidth}{The British Library notes that "Chopin's works have been recorded by all the great pianists of the recording era." \textbf{The earliest recording was an \underline{1895} performance by Paul Pabst of the Nocturne in E major Op. 62 No. 2.} The British Library site makes available a number of historic recordings, including some by Alfred Cortot, Ignaz Friedman, Vladimir Horowitz, Benno Moiseiwitsch, Paderewski, Arthur Rubinstein, Xaver Scharwenka and many others. A select discography of recordings of Chopin works by pianists representing the various pedagogic traditions stemming from Chopin is given by Methuen-Campbell in his work tracing the lineage and character of those traditions.}
& \multirow{6}{\linewidth}{When did Pabst record his Chopin performance?}                           & NQG            & What year was the earliest recording of the earliest recording?                   & 1                               & 2.6                                & 1.6                                   \\
&                                                                                          & BART\textsubscript{LARGE}     & When was the earliest recording of Chopin's Nocturne in E major?                  & 2.8                             & 3                                  & 3                                     \\
&                                                                                          & T5\textsubscript{SMALL}       & When did Paul Pabst perform the Nocturne in E major Op. 62 No. 2?                 & 2.8                             & 3                                  & 3                                     \\
&                                                                                          & T5\textsubscript{LARGE}       & When was the earliest recording of Chopin's Nocturne in E major Op. 62 No. 2?     & 3                             & 3                                  & 3                                     \\
&                                                                                          & -sent-level & When was the earliest recording of Beethoven's Nocturne in E major?               & 1                               & 3                                  & 3                                     \\
&                                                                                          & -answer-free    & In what year did Paul Pabst record Chopin's Nocturne in E major Op. 62 No. 2?     & 3                               & 3                                  & 3                                     \\\midrule
\multirow{6}{\linewidth}{Baird Television Ltd. made Britain's first television broadcast, on 30 September 1929 from its studio in Long Acre, London, via the BBC's London transmitter, using the electromechanical system pioneered by John Logie Baird. This system used a vertically-scanned image of 30 lines – just enough resolution for a close-up of one person, and with a bandwidth low enough to use existing radio transmitters. \textbf{Simultaneous transmission of sound and picture was achieved on \underline{30} March 1930, by using the BBC's new twin transmitter at Brookmans Park.} By late 1930, 30 minutes of morning programmes were broadcast Monday to Friday, and 30 minutes at midnight on Tuesdays and Fridays, after BBC radio went off the air. Baird broadcasts via the BBC continued until June 1932.}
& \multirow{6}{\linewidth}{How many lines made up the picture on Britain's earliest TV broadcasts?} & NQG            & When did the UK's first television broadcast?                                   & 1                               & 2.6                                & 2.2                                   \\
&                                                                                          & BART\textsubscript{LARGE}     & How many lines did Baird Television Ltd.'s first television broadcast use?        & 3                               & 3                                  & 2.6                                   \\
&                                                                                          & T5\textsubscript{SMALL}       & On what date did Baird Television make Britain's first television broadcast?      & 1                               & 3                                  & 3                                     \\
&                                                                                          & T5\textsubscript{LARGE}       & On what date did Baird Television Ltd. make Britain's first television broadcast? & 1                               & 3                                  & 3                                     \\
&                                                                                          & -sent-level & When was Britain's first television broadcast?                                    & 1                               & 3                                  & 3                                     \\
&                                                                                          & -answer-free    & When did Baird Television Ltd. make Britain's first television broadcast?         & 1                               & 3                                  & 2.8                \\\bottomrule                   
\end{tabular}
}
\caption{Examples of the system outputs along with their scores from the manual evaluation. The sentence and answer are highlighted by boldface and underline in the paragraph.}
\label{tab:example-prediction}
\end{table*}

\autoref{tab:example-prediction} presents a few examples of our model predictions with the scores made by the annotators, where the samples are chosen from the high-answerability and low-answerability groups of T5\textsubscript{LARGE}.

\subsection{Spearman's Correlation}
\label{app:correlation}

\begin{figure}[h!]
    \centering
    \includegraphics[width=\columnwidth]{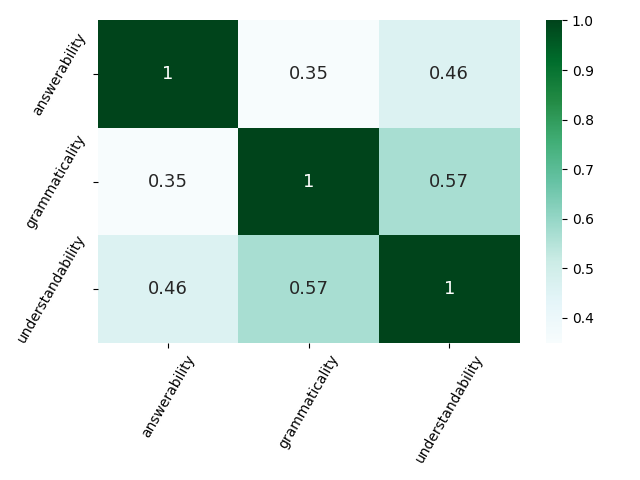}
    \caption{Spearman's rank correlation within manual evaluation criteria. }
    \label{fig:corr_manual}
\end{figure}

\begin{figure}[h!]
    \centering
    \includegraphics[width=\columnwidth]{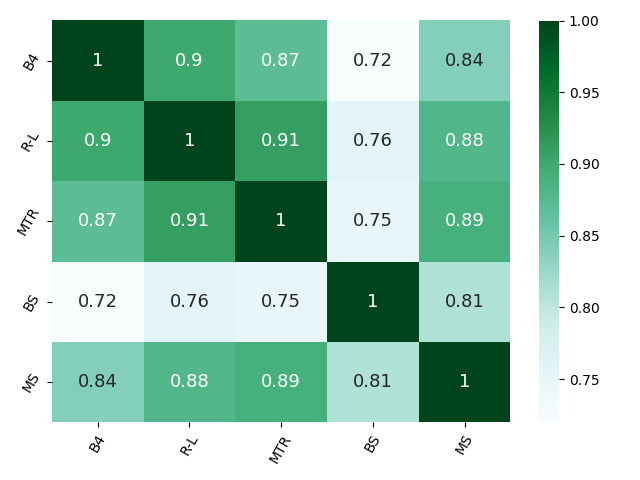}
    \caption{Spearman's rank correlation within automatic evaluation metrics among the 500 samples we used in SQuAD manual annotation.}
    \label{fig:corr_automatic}
\end{figure}

\autoref{fig:corr_manual} and \autoref{fig:corr_automatic} show Spearman's rank correlation across automatic metrics and manual evaluation criteria among the questions we generate over SQuAD test set for the manual annotation. The p-values of all those correlation are less than 0.05 so they are statistically significant.

\subsection{William test}
\label{app:williams}

\begin{figure}[h]
    \centering
    \begin{subfigure}[b]{0.3\textwidth}
        \centering
        \includegraphics[width=\textwidth]{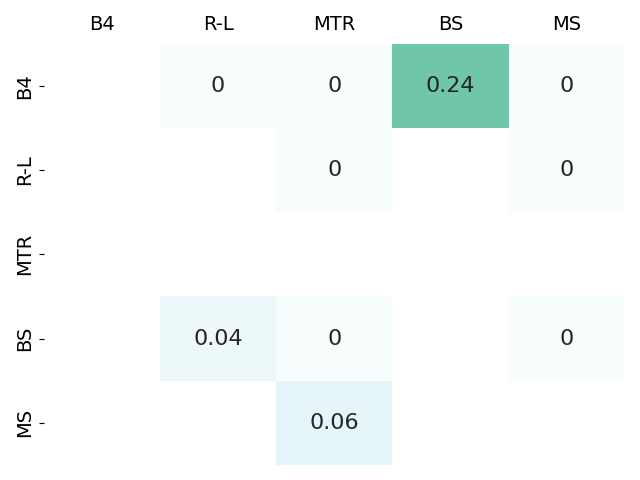}
        \caption{answerability}
    \end{subfigure}
    
    \vspace{5mm}
    
    \begin{subfigure}[b]{0.3\textwidth}
        \centering
        \includegraphics[width=\textwidth]{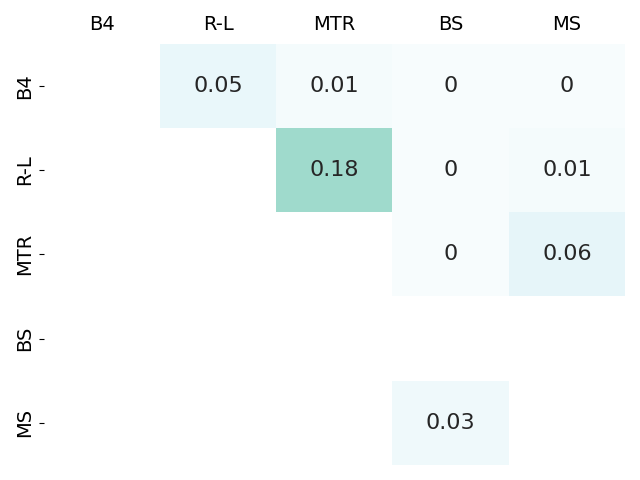}
        \caption{grammaticality}
    \end{subfigure}
    
    \vspace{5mm}
    
    \begin{subfigure}[b]{0.3\textwidth}
        \centering
        \includegraphics[width=\textwidth]{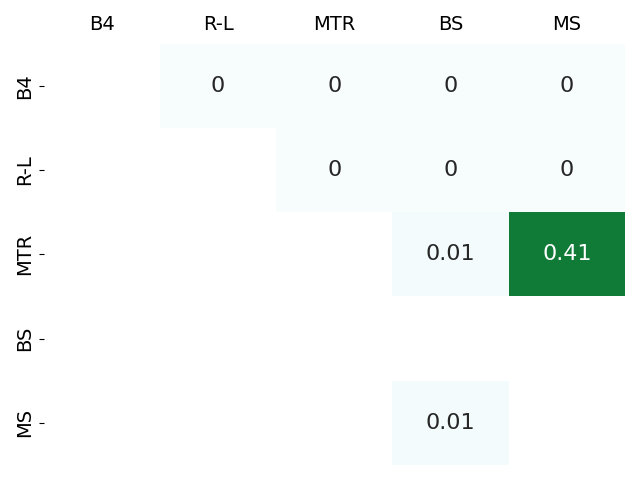}
        \caption{understandability}
    \end{subfigure}
    
    \vspace{5mm}
    
    \caption{Williams test on the difference in the correlation reported in \autoref{fig:corr}. The difference of correlation is significant if the value is less than 0.005.}
    \label{fig:william}
\end{figure}

In \autoref{sec:result_manual_evaluation}, we run correlation analysis and here we report the result of the William test to check the significance of the increase in the correlation across metrics in \autoref{fig:william}, showing that the increase is statistically significant as well.

\subsection{Guidelines}

\begin{figure*}[t]
    \centering
    \includegraphics[width=2\columnwidth]{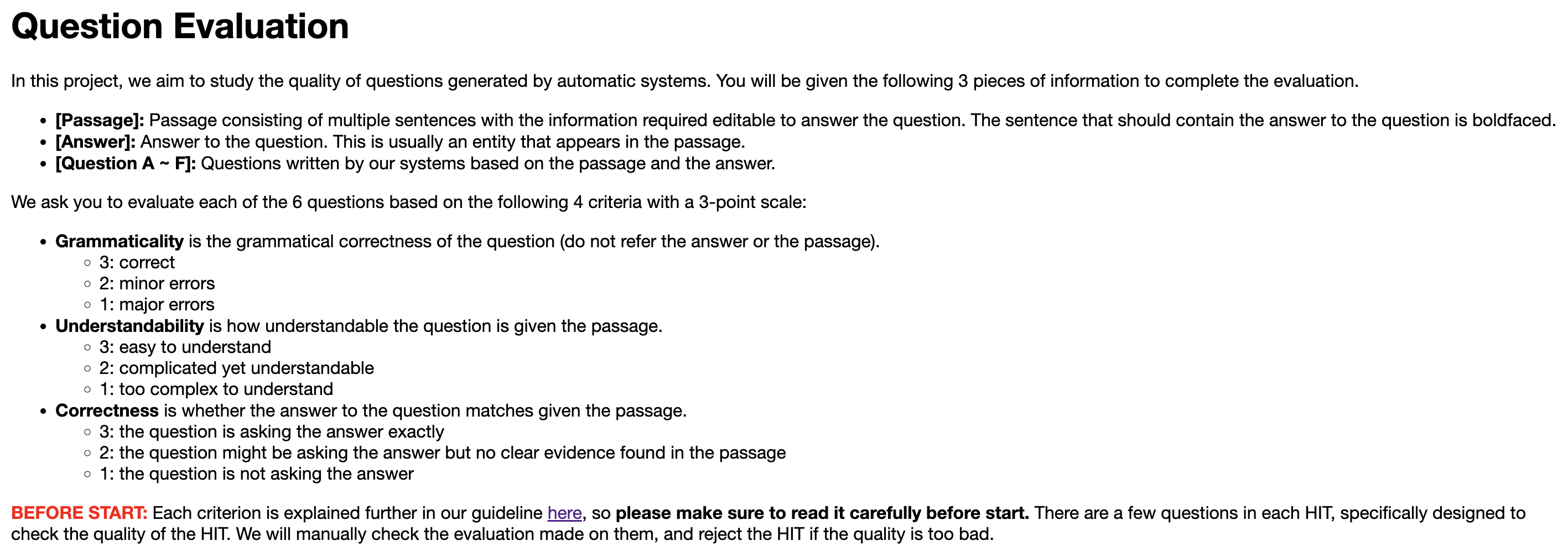}
    \includegraphics[width=2\columnwidth]{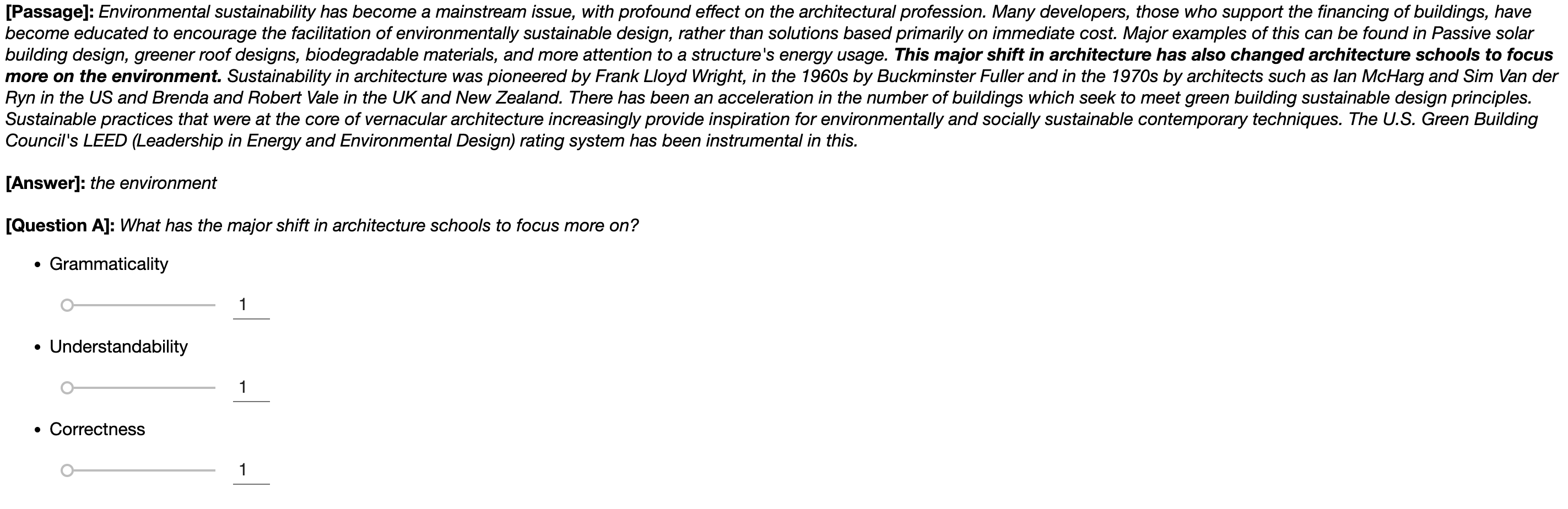}
    \caption{An example of the interface used in our manual evaluation.}
    \label{fig:annotation_interface}
\end{figure*}

\autoref{fig:annotation_interface} shows an example of user interface we implemented for our manual evaluation and the guideline we present to the annotators is attached to the end of the paper.

\section{Unsupervised QA-based Evaluation}
\label{unsupervisedqa}

\begin{table}[t]
\centering
\scalebox{0.75}{
\begin{tabular}{@{\hspace{5pt}}l@{\hspace{8pt}}r@{\hspace{8pt}}r@{\hspace{5pt}}}
\toprule
Data           & F1 & Exact Match \\\midrule
BART\textsubscript{BASE}  & 70.10                  & 58.46                  \\
BART\textsubscript{LARGE} & 70.40                  & 58.60                  \\
T5\textsubscript{SMALL}   & 68.90                  & 56.96                  \\
T5\textsubscript{BASE}    & 70.33                  & 58.14                  \\
T5\textsubscript{LARGE}   & 70.86                  & 59.04 \\\bottomrule
\end{tabular}
}
\caption{Unsupervised QA-based evaluation results of our answer-aware QG models (paragraph-level). All results are the performance on the validation set of original SQuAD by the model trained on the synthetic data generated by each QG model.
}
\label{tab:qae}
\end{table}

As a proxy for \emph{answerability}, we run an unsupervised QA-based evaluation \cite{zhang-bansal-2019-addressing}, which trains a QA model on synthetic data generated by the target QG model and evaluates the QA model on a human annotated test set.
As an alternative to the traditional metrics in QG, Q-metric \cite{nema-khapra-2018-towards} shows high agreement in terms of the \emph{answerability}, but we prefer to employ QA-based evaluation \cite{zhang-bansal-2019-addressing}, since it is more closely tied to downstream applications, while Q-metric relies on some heuristics such as the number of named-entity/pre-defined question types.
This evaluates the QG model's capability to generate high quality questions: higher accuracy of the QA model indicates a better QG model.
The synthetic data is usually generated over the paragraph and answer (PA) pairs collected by \citet{du-cardie-2018-harvesting}.
\citet{zhang-bansal-2019-addressing} used a small subset of the PA pairs, since they contain 12x larger instances than the SQuAD training set. Since this introduces an artifact of the subset choice, we decided to train QA models on the entire PA pairs set with the generated questions.
Also, we train QA models solely on the synthetic data, which differs from work in semi-supervised QA where the QA model is trained on a concatenation of the synthetic data and the original SQuAD training set \cite{lee-etal-2020-generating}.

The synthetic QA data is created by generating a question for each of the one million PA pairs \cite{du-cardie-2018-harvesting} with the target QG model.
We then fine-tune BERT \cite{devlin-etal-2019-bert}\footnote{We use \texttt{bert-base-cased} from HuggingFace.} on the synthetic QA data with the default configuration used in the HuggingFace's tutorial to fine-tune BERT on QA.\footnote{\url{https://github.com/huggingface/transformers/tree/master/examples/pytorch/question-answering}} We report F1 score and the exact match on the SQuAD validation set, following \citet{zhang-bansal-2019-addressing}.\footnote{We will release the synthetic data we made on Huggingface Dataset \url{https://huggingface.co/datasets}.}

The results of our unsupervised QA-based evaluation in \autoref{tab:qae} indicate that the QA model accuracy correlates with the size of QG model that generated the synthetic data, as in T5\textsubscript{LARGE} realizes the best QA model in both of F1 and the exact match,
which is as good as the supervised non-language model based QA models \cite{wang2016machine,yang2017words}. Also, the small models such as T5\textsubscript{SMALL} and BART\textsubscript{BASE} produce QA models with a small decrease in performance, which exhibits the efficiency of our models, similarly to our results with automatic metrics.

\section{Additional Analysis}
\subsection{Zero-shot Multilingual Transfer}

\begin{table}[h]
\centering
\scalebox{0.75}{
\begin{tabular}{lrrrrr}
\toprule
Data  & B4    & R-L   & MTR   & BS    & MS    \\
\midrule
SQuAD & 21.65 & 48.95 & 23.83 & 90.01 & 62.75 \\
\midrule
Ru    & 0.00    & 0.99  & 1.78  & 70.89 & 49.10 \\
Ja    & 0.00    & 6.08  & 0.51  & 66.08 & 46.53 \\
It    & 0.54    & 5.01  & 5.89  & 72.60 & 50.23 \\
Ko    & 0.00    & 0.06  & 0.73  & 66.34 & 45.86 \\
Es    & 0.59    & 5.21  & 6.02  & 74.94 & 50.62 \\
De    & 0.00    & 1.56  & 4.81  & 73.53 & 50.37 \\
Fr    & 1.71    & 15.84 & 8.24  & 72.91 & 50.96 \\ \bottomrule
\end{tabular}
}
\caption{Zero-shot result of mT5 fine-tuned on SQuAD except for the first row, which shows fine-tuning result of SQuAD.}
\label{tab:mlqg-zeroshot}
\end{table}

We fine-tune multilingual language model on each of multilingual QG dataset in \autoref{sec:main_results}, and here we explore the zero-shot multilingual transfer by evaluating English fine-tuned QG model in other languages.
\autoref{tab:mlqg-zeroshot} shows the zero-shot transfer result where we fine-tune mT5\textsubscript{SMALL} on SQuAD and evaluate it on the test set of multilingual QG dataset. 
Compared with \autoref{tab:main-mlqg}, the performance is largely decreased, indicating the difficulty of zero-shot multilingual transfer in QG.

\subsection{Zero-shot Domain Transfer}
\begin{figure}[!h]
    \centering
    \includegraphics[width=\columnwidth]{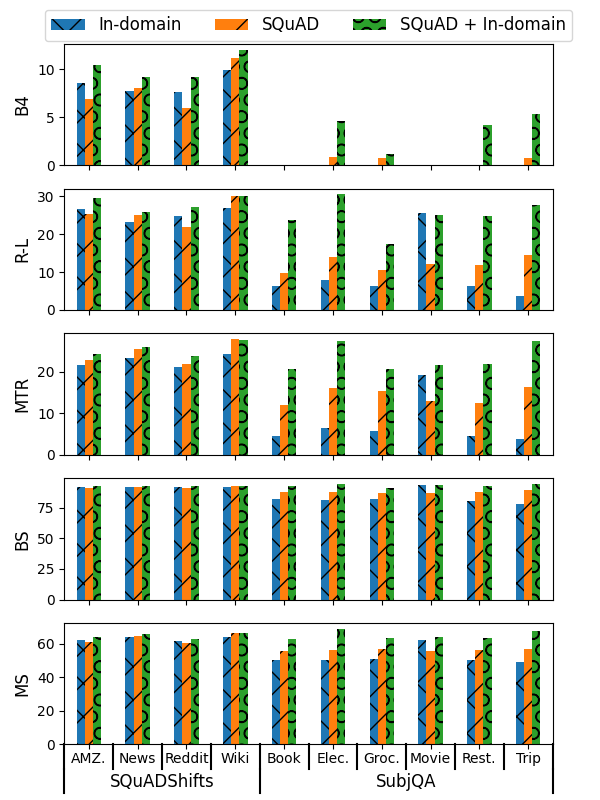}
    \caption{Metric comparison for T5\textsubscript{LARGE} across in-domain fine-tuning, zero-shot transfer of SQuAD fine-tuned model, and in-domain fine-tuning from SQuAD model.}
    \label{fig:domain-difference-t5-large}
\end{figure}

\autoref{fig:domain-difference-t5-large} shows the comparison of zero-shot QG transfer in SQuADShifts and SubjQA dataset with T5\textsubscript{LARGE}.


\clearpage


\section*{Supplementary material}

## Question Evaluation

In this project, we aim to study the quality of questions generated by automatic systems. You will be given the following 3 pieces of information to complete the evaluation.

- **[Passage]:** Passage consisting of multiple sentences with the information required editable to answer the question. The sentence that should contain the answer to the question is boldfaced.
- **[Answer]:** Answer to the question. This is usually an entity that appears in the passage.
- **[Question A ~ F]:** Questions written by our systems based on the passage and the answer.

We ask you to evaluate each of the 6 questions based on the following 4 criteria with a 3-point scale:

- **Grammaticality** is the grammatical correctness of the question (do not refer the answer or the passage).
  - 3: correct
  - 2: minor errors
  - 1: major errors
- **Understandability** is how understandable the question is given the passage.
  - 3: easy to understand
  - 2: complicated yet understandable
  - 1: too complex to understand
- **Correctness** is whether the answer to the question matches given the passage.
  - 3: the question is asking the answer exactly
  - 2: the question might be asking the answer but no clear evidence found in the passage
  - 1: the question is not asking the answer

**BEFORE START:** Each criterion is explained further in our guideline here, so **please make sure to read it carefully before start.** There are a few questions in each HIT, specifically designed to check the quality of the HIT. We will manually check the evaluation made on them, and reject the HIT if the quality is too bad.

**[Passage]:** *Environmental sustainability has become a mainstream issue, with profound effect on the architectural profession. Many developers, those who support the financing of buildings, have become educated to encourage the facilitation of environmentally sustainable design, rather than solutions based primarily on immediate cost. Major examples of this can be found in Passive solar building design, greener roof designs, biodegradable materials, and more attention to a structure's energy usage.* **This major shift in architecture has also changed architecture schools to focus more on the environment.** *Sustainability in architecture was pioneered by Frank Lloyd Wright, in the 1960s by Buckminster Fuller and in the 1970s by architects such as Ian McHarg and Sim Van der Ryn in the US and Brenda and Robert Vale in the UK and New Zealand. There has been an acceleration in the number of buildings which seek to meet green building sustainable design principles. Sustainable practices that were at the core of vernacular architecture increasingly provide inspiration for environmentally and socially sustainable contemporary techniques. The U.S. Green Building Council's LEED (Leadership in Energy and Environmental Design) rating system has been instrumental in this.*

**[Answer]:** *the environment*

**[Question A]:** *What has the major shift in architecture schools to focus more on?*

- Grammaticality

  ○――――― 1

- Understandability

  ○――――― 1

- Correctness

  ○――――― 1

Figure 8: An example of the interface used in our manual evaluation.

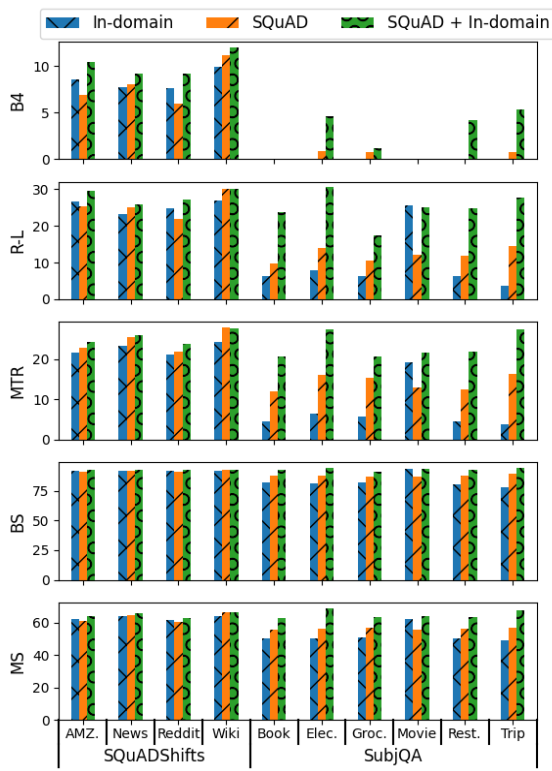

Figure 9: Metric comparison for T5$_{\text{LARGE}}$ across in-domain fine-tuning, zero-shot transfer of SQuAD fine-tuned model, and in-domain fine-tuning from SQuAD model.

# Question Evaluation Guideline

In this project, we aim to study the quality of questions generated by automatic models. You will be given the following 4 pieces of information to complete the evaluation.

- *Passage:* Passage consisting of multiple sentences with the information required to answer the question.
- *Answer:* Answer to the question. This is usually an entity that appears in the passage.
- *Question*: Question written by our model based on the passage and the answer.

## Goal

We ask you to evaluate *questions* based on the following 4 criteria: (1) Grammaticality, (2) Understandability, (3) Correctness, and (4) Question Difficulty.

### (1) Grammaticality

You will score a question in terms of its grammatical correctness with a 3-point scale.
- 3: The question is grammatically correct.
- 2: The question has some minor errors/typos but you can still understand it.
- 1: The question contains many errors and you can not understand it.

You should **only rely on the question** and do not refer to any other information such as the passage and the answer.

### (2) Understandability

You will score how understandable a question is with a 3-point scale.
- 3: The question is easy to read and understand what you are being asked.
- 2: The question is somewhat complicated to understand, but you can get an idea of what the question is asking.
- 1: The question is too complex and you can not understand what should be the answer to the question.

Understandability could correlate with grammaticality, but **a question without any grammatical mistakes can have low understandability** due to an over complex structure. Likewise, a question can be easy to understand even with a few grammatical mistakes. You can refer the passage if it's needed.

### (3) Correctness

We generate a question in a way that its answer should be the answer. Here, you will evaluate whether the answer to The question matches the answer, given the passage.
- 3: The answer to the question is exactly the given answer.
- 2: The answer to the question might be the given answer but no clear evidence can be found in the passage, or the question is too vague. Or the question is not relevant to the passage.
- 1: The answer to the question is not the given answer.

To better understand when you judge a question as 2, let's look at the following example:
- Passage: Max was raised in LA …
- Answer: LA
- Question: Where was Max born?

While the answer "LA" could be an answer to the question, it is not entirely accurate **since the passage does not explicitly state that Max was born in LA**. A more correct question could be "Where was Max raised?". In these situations, you could score the question with a 2.

The answer can sometimes be partial but it is fine. The answer in the following example should be `June 6th 1992` instead of `June` but you should score it with a 3 as the question still makes sense with the answer June.
- Passage: Max was born on June 6th 1992, and …
- Answer: June
- Question: When was Max born?

In some cases, the question matches the answer while it is completely irrelevant to the passage. For example,
- Passage: China spans five geographical time zones and borders 14 different countries, the second most of any country in the world after Russia.
- Answer: China
- Question: What is the name of the world's most populous country?

Although the question matches the answer, it is based on common knowledge rather than any evidence found in the passage, so this question should be scored as 2.

In addition, if the Understandability of the question is 1, you should mark its Correctness as 1.




\end{document}